**Title:** "Evolving soft locomotion in aquatic and terrestrial environments: effects of material properties and environmental transitions"

**Running title (~45 chars):** Evolving soft locomotion

**Authors names and affiliations:**

Francesco Corucci[1,2,*], Nick Cheney[2,3], Francesco Giorgio-Serchi[4], Josh Bongard[2] and Cecilia Laschi[1]

[1] The BioRobotics Institute, Scuola Superiore Sant'Anna, Pisa, Italy
[2] Morphology, Evolution & Cognition Lab, University of Vermont, Burlington, VT, USA
[3] Department of Biological Statistics and Computational Biology, Cornell University, Ithaca, NY, USA
[4] Fluid Structure Interaction Research Group, Southampton Marine and Maritime Institute, University of Southampton, Southampton, UK

* **Correspondence:** f.corucci@santannapisa.it

## Abstract (<250 words):

Designing soft robots poses considerable challenges: automated design approaches may be particularly appealing in this field, as they promise to optimize complex multi-material machines with very little or no human intervention. Evolutionary soft robotics is concerned with the application of optimization algorithms inspired by natural evolution in order to let soft robots (both morphologies and controllers) spontaneously evolve within physically-realistic simulated environments, figuring out how to satisfy a set of objectives defined by human designers. In this paper a powerful evolutionary system is put in place in order to perform a broad investigation on the free-form evolution of walking and swimming soft robots in different environments. Three sets of experiments are reported, tackling different aspects of the evolution of soft locomotion. The first two sets explore the effects of different material properties on the evolution of terrestrial and aquatic soft locomotion: particularly, we show how different materials lead to the evolution of different morphologies, behaviors, and energy-performance tradeoffs. It is found that within our simplified physics world stiffer robots evolve more sophisticated and effective gaits and morphologies on land, while softer ones tend to perform better in water. The third set of experiments starts investigating the effect and potential benefits of major environmental transitions (land ↔ water) during evolution. Results provide interesting morphological exaptation phenomena, and point out a potential asymmetry between land → water and water → land transitions: while the first type of transition appears to be detrimental, the second one seems to have some beneficial effects.

# 1 Objective

Designing soft robots [1] [2] [3] represents a considerable challenge. The design space of these machines is vast and complex, and their effective behavior often arises from complex interactions among controller, morphology, and environment, which are not trivial to foresee. This usually results in the necessity to design, fabricate, and test multiple designs, which requires time and resources. In order to alleviate this problem, effective and realistic physical simulation tools have been developed in the past years [4] [5], which allow to analyze different robot configurations before fabricating and testing them in the real world. Although transferring a design from simulation to reality is not always straightforward [6], computer simulations can provide useful information and save time and resources before attempting tests in the real world. Moreover, increasingly sophisticated techniques that allow to preserve the effectiveness of simulated design once deployed in the real world are being developed [7] [8].

In addition to leveraging simulation tools to manually design and analyze soft robots prior to their fabrication, particularly appealing is the possibility to automatize their design altogether, with algorithms capable of automatically discovering effective morphologies and controllers for a given task and environment. This is what is done in fields such as *evolutionary robotics* (or *evo-robo*) [9], where researchers develop effective ways to encode complete robot designs [10] that are then automatically optimized thanks to powerful optimization algorithms (*evolutionary algorithms* [11] [12]). By abstracting certain properties of natural evolution, these algorithms aim at instantiating evolutionary dynamics in computer simulations (*"in silico"*), hoping to automatically produce a diversity of effective and well adapted solutions that may, one day, approach the complexity and sophistication of those found in the biological world. These techniques have recently been applied in soft robotics as well: in the field of *evolutionary developmental soft robotics* (*evo-devo-soro*) [13] [14] [15] computational processes inspired by biological evolution and development are put in place in order to let soft robots *grow* and *evolve* [16] artificial brains and multi-material bodies spontaneously, within physically-realistic simulations, letting them figure out on their own how to satisfy a set of objectives provided by the designer. In light of this biological inspiration, evolutionary and developmental simulations can not only be used for engineering purposes (i.e. to devise optimized and adaptive robot designs able to solve certain tasks), but can also be interpreted with a more biological flavor, interpreted as instances of life-like processes, as done in the *artificial life* [17] field.

The authors have applied these techniques in several domains: some examples regard solving complex engineering problems associated with soft robotics (parameters identification) [18] [19], discovering effective gaits and morphological configurations for an existing soft robot [20] [21] [22], as well as exploring properties such as embodied intelligence [23] [24] and morphological computation [25] [26] [27] [28] [29] in evolved soft robots [30] [31].

Taking the cue from [32], in this paper we are interested in performing a broad investigation on the free-form evolution of soft robots performing locomotion in different environments, i.e. in aquatic and terrestrial ones. Thanks to a general evolutionary system (composed of an effective soft robot simulator, a powerful generative encoding and a multi-objective evolutionary algorithm), we are interested in observing how different material properties affect the evolution of different morphologies, behaviors, and energy-performances tradeoffs, in different environments. In addition to showcasing the potential of evolutionary techniques to produce a wide array of well adapted walking and swimming soft creatures, our analyses may provide new insights regarding the evolution of soft locomotion.

Several works have dealt with different aspects of the evolution of walking and swimming robots in the past. Nevertheless, very few have extensively compared the free-form evolution of soft creatures in both environments, and analyzed the role of material properties in such evolution. Ijspeert et al. have focused on the evolution of neurologically-inspired controllers for fixed morphologies, as well as on the swimming → walking transitions from a control point of view [33] [34] [35] [36]. Similar approaches, mostly focused on the evolution of neural control systems, are reported in [37] [38] [39]. Some robotics works have focused on evolving gaits for fixed soft robots [40]. Others have evolved control and stiffness properties for fixed morphologies embedding flexible elements [41] [42], which were in some cases experimentally validated. As for the free-form evolution of virtual creatures, robots have been evolved both in water and on land since the very first attempt from Karl Sims [43], where evolved creatures were, however, made by interconnected rigid elements. More recently researchers have focused on evolving locomotion in rigid-bodied creatures actuated by flexible muscle-like systems [44] [45] [46], which represents an additional step towards the evolution of more lifelike creatures and robots. Contributions to the free-form evolution of terrestrial soft creatures come from Rieffel et al. [16] [47] [48] [49] [50], as well as Hiller et al. [51] [52], Cheney et al. [53] [54] [55] and Methenitis et al. [56]. Closely related is also the work by Joachimczak et al., which have evolved both artificial brains and bodies for multicellular soft robots, both in fluid and terrestrial environments [57] [58] [59], and have even attempted transitions between the two by exploiting the concept of *metamorphosis* [60].

In what follows we will start by describing the evolutionary system we are going to use for our experiments, consisting of a soft robot simulator (Sect. 2.1) that we augment with a simple fluid model (Sect. 2.2), a powerful generative encoding (Sect. 2.3), and a multi-objective evolutionary algorithm (Sect. 2.4). Experiments (Sect. 2.5) will be then reported (Sect. 3) in which robots characterized by different material stiffness are evolved, both on land (Sect. 3.1) and in water (Sect. 3.2). Finally, we will report preliminary attempts aiming at investigating the effect and potential benefits of environmental transitions water ↔ land on the evolution of soft morphologies and behavior (Sect. 3.3). Widely studied by biologists and paleontologists with the aim of finding evolutionary links between aquatic and terrestrial species [61] [62] [63] [64] [65] [66] [67] [68] [69] or understanding evolutionary pressures which determine such transitions [70] [71] [72] [73], the study of adaptations associated with major environmental changes may provide interesting insights to the field of evolutionary robotics and artificial life as well.

## 2 Materials and Methods

### 2.1 Physics engine

The simulation environment here adopted is based on the VoxCAD simulator [4], an open source C++ voxel editing software which includes a graphical interface as well as an efficient physics engine (Voxelyze). VoxCAD allows the quantitatively accurate simulation [52] of the static and dynamic behavior of free-form 3D multi-material structures, which can be characterized by large deformations and heterogeneous properties (e.g. density, stiffness). The physics simulation is based on a lattice of voxels interconnected by flexible beams, which is used by Voxelyze in order to compute the dynamics of the soft body. A cubic (or rectangular) workspace can be populated by an arbitrary number of interconnected voxels. A deformable mesh representing the robot is rendered as well in VoxCad, which usually only serves for visualization. Nevertheless, in order to empower the simulator with a fluid environment, the computation of fluid dynamics forces will be based on the interaction of this deformable mesh with the surrounding fluid. Resulting forces will be then applied to the underlying voxel based representation. In order to facilitate our experiments, a high-level Python library was developed [74], which implements state of the art evolutionary algorithms and generative encodings. This Python wrapper invokes Voxelyze in order to perform fitness evaluation.

### 2.2 Simulated environments

In what follows simulated robots will be evaluated in two different environments, a terrestrial and an aquatic one. The terrestrial environment is provided by the off-the-shelf VoxCad simulator, and consists of a smooth and flat ground. As an aquatic environment is not provided by the simulator, and was thus implemented in order to support our experiments. The fluid model consists of a simplified mesh-based quadratic drag model (Figure 1).

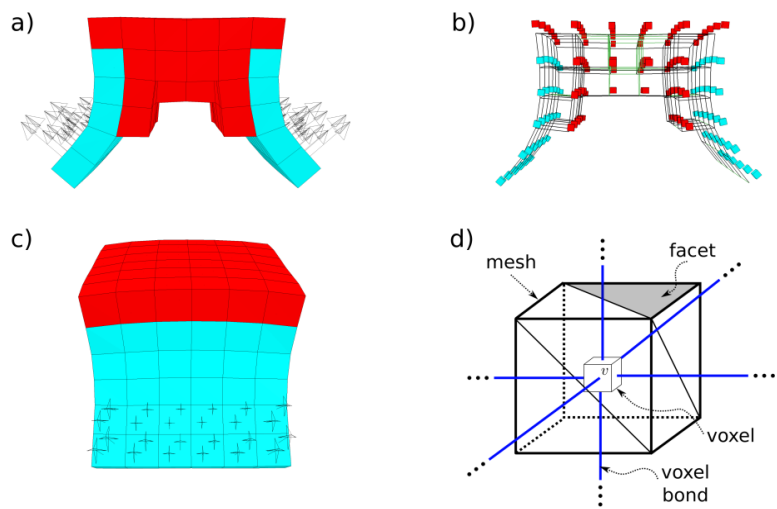

Figure 1 Fluid model. Drag forces are computed on the deformable mesh representing the robot (a) and applied to the underlying lattice of voxels (b), which is used by Voxelyze to compute the dynamics of the body. c) Lateral view: note how two vectors depart from each voxel currently subject to noticeable drag forces. This is due to each superficial square facet visible in the picture being actually composed of two triangular facets, each one contributing to the drag computation. A voxel with no neighbors can receive drag forces from all 12 surrounding facets exposed to the fluid. Voxels with at least one neighbor will receive drag contributions from a lower number of facets. Drag forces visible in a) and c) come from the inner surface of this flapping creature, which is moving upwards.

A local drag force is computed for each facet of the deformable mesh associated with each robot, and added up as an additional force experienced by the underlying voxels. The total drag force $\vec{F}_{dv}$ experienced by a voxel $v$ is:

$$\vec{F}_{dv} = \sum_{i=0}^{N} \vec{F}_{dfi}$$

where $N$ is the number of facets surrounding the voxel $v$ and $\vec{F}_{dfi}$ is the drag force experienced by the $i$-th facet:

$$\vec{F}_{dfi} = -\frac{1}{2} \rho_f\, C_d\, A_f\, \vec{v}_f^{\,2}$$

where:

- $\rho_f$ is the fluid density;
- $C_d$ is the facet's drag coefficient (set to 1.5);
- $A_f$ is the area of the (possibly deformed) facet;
- $\vec{v}_f$ is the facet's normal speed, achieved by projecting the speed of the corresponding voxel along the facet's normal.

By assuming neutral buoyancy, the fluid model is completed by simply setting the gravity acceleration to zero. The fluid dynamics model is extremely simplified in order to reduce the computational costs associated with fitness evaluation. Robots will evolve in an environment where the only fluid dynamics force is the resistive drag, and no turbulent phenomena exist. Particularly, the approximations employed here neglect two components that are very relevant to locomotion in a fluid environment [75], consisting of inertial and viscous fluid terms. The former are associated with the reactive forces due to the acceleration of a body (or part of it) in a fluid, and partly characterize the response of the surrounding medium to the beating of a tail, the oscillation of a fin or the forced expulsion of a slug of fluid. The latter are the result of the shedding of vortices from the boundary layer which forms over the body (or parts of it), which contribute to forces oriented either in the tangential or normal direction with respect to the body of interest, thus generating drag in one case, and lift/thrust, in the other.

As we are well aware of the importance of these phenomena for animal and robot locomotion in real aquatic environments, we will limit the scope of our results to our simulated environment.

## 2.3 Genetic representation/encoding

### 2.3.1 Compositional Pattern Producing Networks (CPPNs)

The genetic representation is based on a very general and evolvable network encoding known as Compositional Pattern Producing Networks (CPPNs) [76], which is used in combination with the VoxCad simulator as in a number of previous studies [51] [53] [55] [54] [56]. Designed to capture the formation of regular patterns during development without modeling developmental dynamics per se [77], CPPNs encode the genetic material as a composition of functions, represented in the form of a graph, or network. Each network is composed of a variable number of interconnected nodes, each characterized by an activation function. The number of nodes, the specific activation functions, links between nodes and weights of these links can all be placed under evolutionary control and evolved by means of algorithms such as CPPN-NEAT [76].

In CPPNs *"a function produces a phenotype"* [76], the function being implemented by the particular set of nodes and connections composing the CPPN. The input to the network usually consist in spatial information (e.g. the location of a cell) that, convolved through the network, result in outputs that are interpreted as phenotypic traits (e.g. the type of a cell at that particular location).

A complete phenotype is achieved by querying the evolved CPPN at every location of the workspace: as neighboring cells have similar spatial information (coordinates), they also tend to produce similar outputs from the networks, creating a bias towards continuous and spatially organized patches of given material, as found in biological organisms. An additional bias towards regular phenotypic patterns is due to the set of activation functions which are allowed at each node, usually chosen from a pool of regular functions such as sigmoids, linear functions, sines, gaussians, etc.

By design, this representation is particularly useful in order to encode spatially distributed properties such the parameters associated with distributed sensory and control systems. Moreover, both CPPN inputs and outputs are real-valued, which means that the system automatically supports continuous phenotypes of arbitrarily resolution. In our experiments some phenotypic traits will be continuous, while in some other cases continuous CPPN outputs will be translated in a number of discrete values by means of thresholding and sequential expression patterns, as done, e.g., in [53].

In our setup [74] the experimenter can decide how to structure the genetic material (inputs, outputs, and number of evolved CPPNs), and how to interpret CPPNs outputs in order to build each phenotypic trait. A single CPPN with several outputs can be evolved: this promotes the reuse of genotypic information, as shared substructures of the same CPPN can contribute to different phenotypic traits. However, monolithic CPPNs are more prone to pleiotropic effects (i.e. when a mutation on a gene affects two or more unrelated phenotypic traits), which can be detrimental in some cases. An alternative choice is to evolve several independent networks, if genotypic modularity is deemed to be enforced.

### 2.3.2 Structuring the genetic material

The genetic material of the virtual creatures evolved in this chapter is structured in two different CPPNs: one encodes phenotypic traits associated with the morphology, another encodes those associated with the control (Figure 2). This is not an arbitrary choice, and is actually based on ongoing research on some aspects related to the co-evolution of artificial brains and bodies [78], which suggest how morphology and control may have a different role in brain-body co-evolution, thus justifying handling them separately.

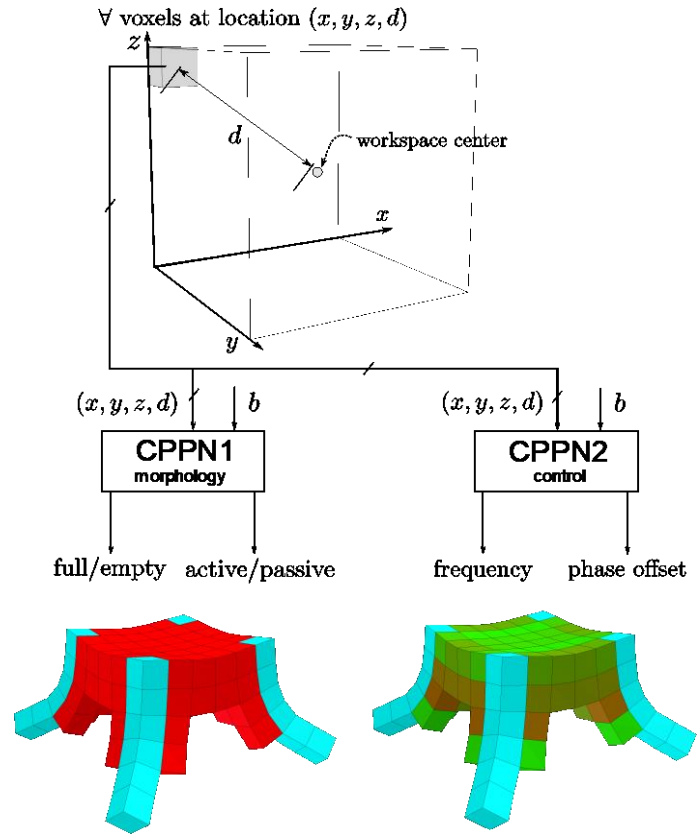

Figure 2 Structure and expression of the genetic material. Different aspects of the phenotype are color coded and shown in two different views of the same robot. Red voxels are active (actuated), cyan ones are passive. The green gradient on the right represents the actuation state of each active voxel, which depends on the outputs of CPPN2.

For every voxel, both CPPNS receive the same four spatial inputs $(x, y, z, d, b)$, where $(x, y, z)$ are the cartesian coordinates, $d$ is the distance of the voxel from the center of the workspace, $b$ is a bias.

The morphological CPPN has one output which dictates whether a voxel is full (output $\geq 0$) or empty (output $< 0$). If a voxel is full, the second output is examined, which dictates whether the voxel is *active* (i.e. actuated, output $\geq 0$, displayed in red in Figure 2) or *passive* (i.e. non-actuated, output $< 0$, displayed in cyan in Figure 2). This is an example of a sequential expression pattern translating continuous outputs into discrete phenotypic traits.

Actuation is realized through a global sinusoidal oscillation, which controls the volumetric expansion/contraction of each voxel: its frequency is achieved by averaging the first output of the control CPPN (*frequency*) output of all voxels[1], with a *phase offset* which can be different for each voxel depending on the second output. In this case one output (*frequency*) is aggregated into a global quantity, while the second one (*phase offset*) is treated as a continuous output.

---

[1] This was done for simplicity. It would be interesting to remove this simplification and allow each voxel to contract at a different frequency, which can be easily achieved.

## 2.4 Artificial Evolution

As it is common in the field of *evolutionary robotics* [9], in this work robots are optimized by means of *evolutionary algorithms* (or EAs) [11] [12], which are population-based metaheuristic optimization algorithms inspired by natural evolution. EAs manipulate populations of candidate solutions (or *individuals*, representing different robot designs in this setting) mimicking processes of natural *reproduction, mutation, recombination*, and *selection*. Overall, they try to capture the *algorithmic nature of natural evolution, which produces innovation through the non-random selection of random variations*. The process starts from a randomly initialized population of diverse individuals, whose characteristics (*phenotypic traits*) are encoded in a *genotype*, which is represented by a data structure (e.g. a vector, a network, etc.). The population of simulated robots evolves over a number of *generations* (iterations of the algorithm): at each generation the current population is evaluated in the *task environment* at hand. The fitness of each individual is assessed by a function (*fitness function*) which assigns one scalar number (or more, in *multi-objective* [79] settings) to each individual. Individuals that perform better with respect to the task at hand get higher chances of surviving to the next generation and reproducing, breeding offspring that consist in slightly modified copies of themselves. During this process, each selected individual can be altered by evolution through *genetic operators*, which operate on the genotype. These can mimic both genetic *mutations* (i.e. random perturbations of the genetic material of an individual) as well as genetic *recombinations*, where the genetic material of two parents is combined to form one or more children.

In what follows, a multi-objective evolutionary algorithm is adopted [2], which allows to optimize several objectives at the same time based on the concept of *pareto dominance* [79]. The algorithm only exploits genetic mutations.

The following three objectives were defined:

1. Maximize the distance traveled by the center of mass of the robot in the allocated time[3];
2. Minimize the percentage of actuated voxels (energy);
3. Minimize the number of voxels (employed material).

The first objective selects for locomotion. The second one favors robots which use less energy to move around, which forces evolution towards a parsimonious use of energy, with the aim of promoting the evolution of energy-efficient trade-offs between performances and energy usage. The third one pushes evolution towards a parsimonious use of material, with the aim of favoring the evolution of interesting morphologies.

### 2.4.1 Shape descriptors

As we are interested in characterizing different properties and regularities of the evolving robots, we have designed and/or implemented a number of shape descriptors, described in what follow.

---

[2] An implementation is available at https://github.com/skriegman/evosoro
[3] The allocated time consists in a fixed number of actuation cycles, and can thus vary for robots characterized by different actuation frequencies.

### 2.4.1.1 Symmetry

In order to capture different types of morphological symmetries (such as reflection and translation symmetries along various axes and directions) we introduce a *global symmetry index* ($G_{SI}$).

The descriptor is computed as follows (Figure 3, Figure 4):

1. The bounding box of the robot is computed;
2. The robot is then analyzed along the $x, y, z$ axes, separately;
3. Each 2D slice $s_i$ of the robot encountered while analyzing it in one of the three possible axes is sequentially analyzed;
4. A symmetry index is computed for each type of symmetry being considered, which include four possible reflection symmetries (vertical, horizontal, along the primary and secondary diagonal) and two possible translation symmetries (vertical and horizontal). A total of six indices is thus computed, respectively: $r_{v_i}, r_{h_i}, r_{pd_i}, r_{sd_i}, t_{v_i}, t_{h_i}$. These are computed by applying the relevant transformation to the robot slice, and computing the percentage of overlapping voxels (result $\in [0,1]$) (Figure 4);
5. Indices from all 2D slices encountered along one axis are aggregated through the mean, e.g. $r_{v_x} = \text{mean}(r_{v_{i_x}})$. This yields six different descriptors capturing symmetries along each of the three axes $x, y, z$;
6. These indices are further aggregated in order to achieve a single symmetry descriptor capturing overall symmetry along a given axis. As an example, the symmetry index along the $x$ will be $s_x = \text{mean}(r_{v_x}, r_{h_x}, r_{pd_x}, r_{sd_x}, t_{v_x}, t_{h_x})$. Descriptors along the other two axes ($s_y, s_z$) are computed analogously. This yields three symmetry descriptors $s_x, s_y, s_z \in [0,1]$, which capture different types of symmetries along each of the three axes $x, y, z$;
7. Finally, a global symmetry index can be computed by further aggregating the three symmetry indices $s_x, s_y, s_z$, as follows: $G_{SI} = \text{mean}(s_x, s_y, s_z) \in [0,1]$.

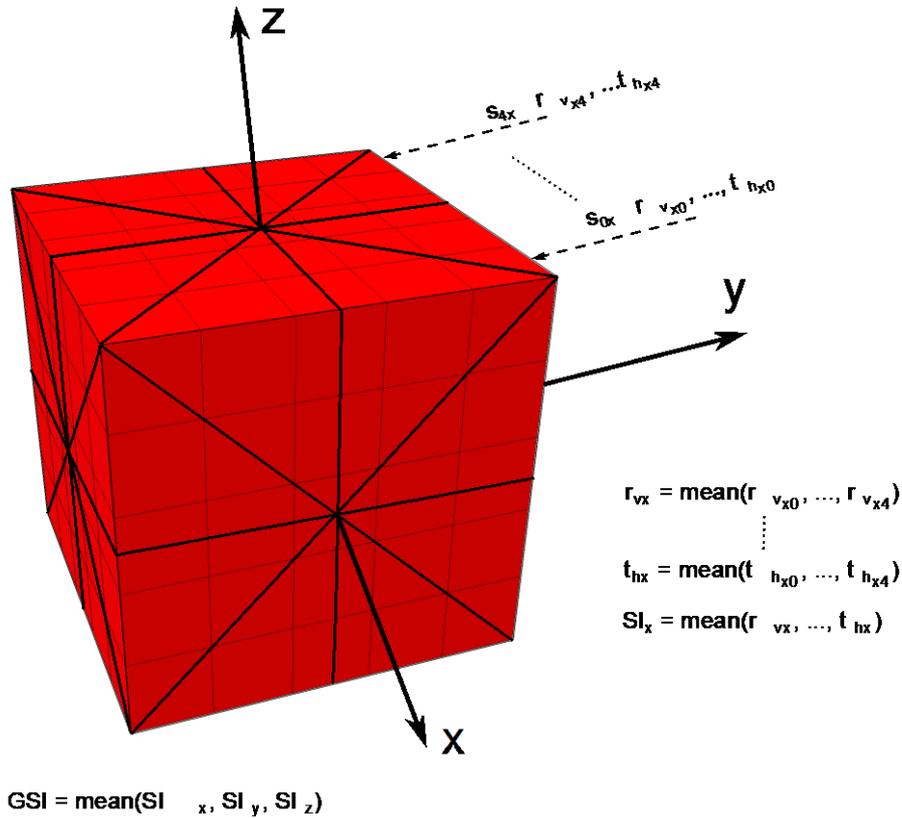

Figure 3 Symmetry descriptors, as described in the text. The robot is navigated along the three main axes $x, y, z$, and six different descriptors are being computed on each encountered 2D slice, capturing different types of reflection and translation symmetries. Each of these six descriptors is aggregated for each slice encountered along a given axis, and further aggregated yielding a single global symmetry index capturing the symmetry of the whole robot.

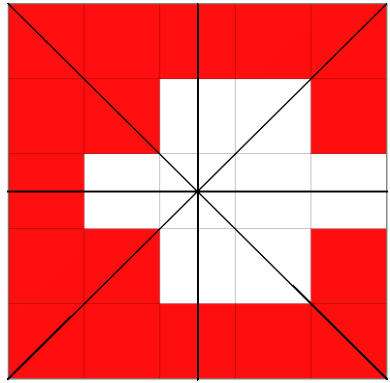

Figure 4 Example of symmetry descriptors computed on a 2D slice of the robot. $(r_{v_i}, r_{h_i}, r_{pd_i}, r_{sd_i}, t_{v_i}, t_{h_i}) = (0.7, 1.0, 0.8, 0.8, 0.7, 0.6)$

### 2.4.1.2 Branching

A *branching index* (BI) is introduced in order to capture the emergence of hollow structures as well as appendages such as legs in the evolving robots. This descriptor is computed as follows:

$$BI = 1 - \frac{R_v}{CH_v} \in [0,1]$$

Where $R_v$ is the robot volume (computed with a mesh-based method [80]) and $CH_v$ is the volume of its *convex hull* (computed as in [81]), i.e. the smallest convex volume that encloses the robot. A higher BI denotes a more "structured" morphology, while a low one corresponds to less interesting shapes such as full cubes (BI = 0). An example is reported in Figure 5.

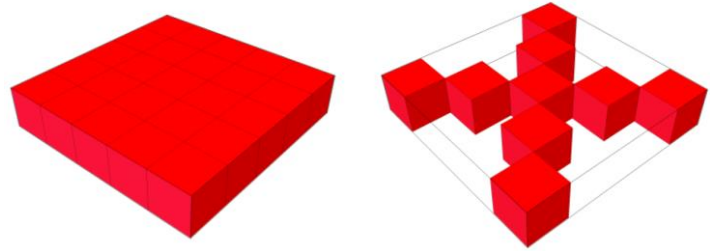

Figure 5 Branching index for two different robots, example. The one on the left has $BI = 1 - \frac{25}{25} = 0$, the one on the right has $BI = 1 - \frac{9}{25} = 0.64$. The higher BI captures a more complex morphological structure in terms of appendages and volume distribution.

### 2.4.1.3 Morphological complexity

Another useful descriptor here adopted is a metric of *morphological complexity*. As in [82], morphological complexity is here quantified in information theoretic terms as the *entropy of curvature* (here also referred to as *shape entropy*), which is computed on the undeformed mesh representing the robot. This metric, based on Shannon's entropy [83], is computed by considering the local curvature of the mesh representing the robot as a random variable: is has been shown [84] [85] that the entropy of such a variable quantitatively and qualitatively captures a notion of morphological complexity which agrees with the intuitive understanding of humans. Intuitively, in order to have higher complexity, a morphology must exhibit many angular regions (i.e. regions of non-zero curvature), and those angles should not follow a predictable, repeating pattern.

## 2.5 Experiments

Various sets of experiments have been performed (Table 1). In some of these, robots evolve in a static environment (either water, or land) for 5000 generations. These experiments have been performed in five different conditions, differing in the stiffness (elastic modulus) of the single material provided to evolution, which varied from S1 = 0.001 MPa (softest) to S5 = 10 MPa (stiffest), spanning four orders of magnitude (Table 2). This first batch of experiments represents the core of this paper.

Additionally, two other sets of experiments have been performed, in which robots evolve for a total of 9000 generations, spending the first 4500 in one environment (water/land), before the entire population is moved to the second environment (land/water) for further evolution (for another 4500 generations). For simplicity, the transition from one environment to the other was abrupt, although more gradual shaping techniques would be desirable and may be implemented in future work (e.g. gradually introducing robots to the new environment, either by implementing compound land-water environments, or by gradually altering the physics). Due to computational constraints, environmental transition experiments were performed for a single, intermediate value of the material stiffness (S3). Among the runs characterized by static environments, S3 runs were allowed to continue for 9000 total generations instead of 5000, in such a way to later perform a fair comparison (same number of generations) among these robots with those evolved in presence of environmental transitions.

| Stiffness | Value [MPa] |
|---|---|
| S1 | 0.001 |
| S2 | 0.01 |
| S3 | 0.1 |
| S4 | 1 |
| S5 | 10 |

Table 1 Stiffness of the material evolution is provided with in different experiments.

| Environment | Material Stiffness | Total generations |
|---|---|---|
| Land | S1, S2, S3, S4, S5 | 5000 |
| Water | S1, S2, S3, S4, S5 | 5000 |
| Land → Water (halfway) | S3 | 9000 |
| Water → Land (halfway) | S3 | 9000 |

Table 2 Summary of the experiments reported in the paper

For each treatment, 15 repetitions were performed, with different seeds being provided to the random number generator. As for the statistical analysis of the results presented in what follows, confidence intervals reported in the plots were estimated at a 95% level using a bootstrapping method. P-values were estimated with the Mann-Whitney U test, and the Bonferroni correction for multiple comparisons was applied whenever necessary.

## 3 Results

### 3.1 Evolution on land

Figure 6 reports the fitness curves for the five experiments with different material stiffness, while Figure 7 compares a number of descriptors computed on the final populations.

From Figure 7 a clear increasing trend for fitness can be observed as we move from the softer setting towards stiffer ones (S4). The increasing trend stops after S4, where fitness seems to saturate (if not slightly decreasing, although not statistically significant – n.s.). This may suggest that the S4 stiffness is optimal for terrestrial locomotion, at least in our simulated environment, and for the range of stiffness values here explored. This result underscores the correlation between body stiffness and gravity in the frame of land-based locomotion, whose trend is further highlighted in Sect. 3.2. Simple scaling analysis shows that, for a body of average density similar to that of rubber and subject to a gravitational force analogous to that experienced on Earth, the elastic modulus of S4 causes a deformation of about 1.5%. This is indeed a suitably low degree of deformation for legged locomotion to develop. At S5, the deformation suffered by a limb in the simulated conditions is as low as 0.15%, not providing any additional feature to those already evolved in the S4 case, as confirmed in

Figure 8 (lower subset). On the other hand, at the stiffness values of S3 and lower, deformation can attain values from 15% upwards, impeding the development of structures capable of sustaining the body over the ground and thus preventing limb-supported mobility, see Fig. 8 (lower subset).

To sum up, in our setting, the stiffer the robots, the better the locomotion performances.

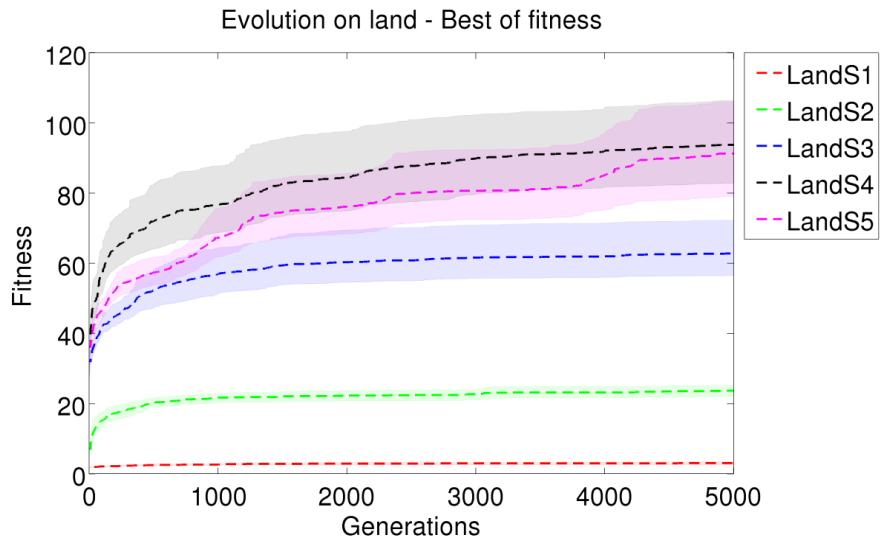

Figure 6 On land, stiffer robots evolve faster locomotion. See Figure 7 for a clear statistical comparison among the five treatments. Note how the softest treatment (S1) corresponds to non-functional robots.

Note how robots which evolved on land for stiffness S1 have a fitness which is very close to zero. These robots are not functional, and the only fitness they are able to score is due to them sprawling and unfolding onto the ground, as will be clear from the qualitative analysis of evolved behaviors (Figure 8). The softest material (S1) results to be too soft to support any form of terrestrial locomotion. In addition to being consistent with biological observations (very soft aquatic creatures -- like jellyfishes -- are not able to move on land, collapsing under the effect of gravity), this observation is important as it may entail some biases in the trends we will observe in what follows. We are interested in trends and regularities related to locomotion: if locomotion cannot be evolved at all (as in S1), all other descriptors and their comparison with those computed in the other treatments risk to be not meaningful. This is particularly true in light of the multi-objective optimization setup. Locomotion is but one of three optimization objectives. If all robots are poor in the same way in the covered distance, evolution becomes free from the constraint of producing functional robots, and will try optimizing the remaining objectives. For example, it may add a lot of passive material to the non-functional robots, which may be mistaken for the existence of highly energy efficient locomotion strategies in S1, while in fact all robot fail to move at all. The same goes for other descriptors.

Figure 7 reveals other interesting information. In addition to moving further, stiffer robots also use less energy to move around (with the only exception of S1, on which we have already commented). This might be due to the fact that on land, as robots get stiffer, they become better able to sustain more efficient locomotion strategies (e.g. walking, as opposed to crawling/wiggling around/vibrating). Qualitative observations seem to agree with this hypothesis (Figure 8).

The increasing trend of the branching index also confirms these observations: stiffer robots are able to develop more complex appendages, lifting up from the ground and achieving more effective forms of terrestrial locomotion. The decreasing trend observed in the actuation frequency also points in this direction, with slower, coordinated walking gaits opposed to fast robots which move by vibrating or bouncing. Overall, higher stiffness seems to be more adequate for terrestrial locomotion, entailing better performances, less energy consumption, and evolution of more elaborate morphologies and behaviors (Figure 7). This is also evident from the analysis of the aggregate pareto fronts (Figure 8), which shows how stiffer robots achieve better trade-offs between locomotion performances and energy usage.

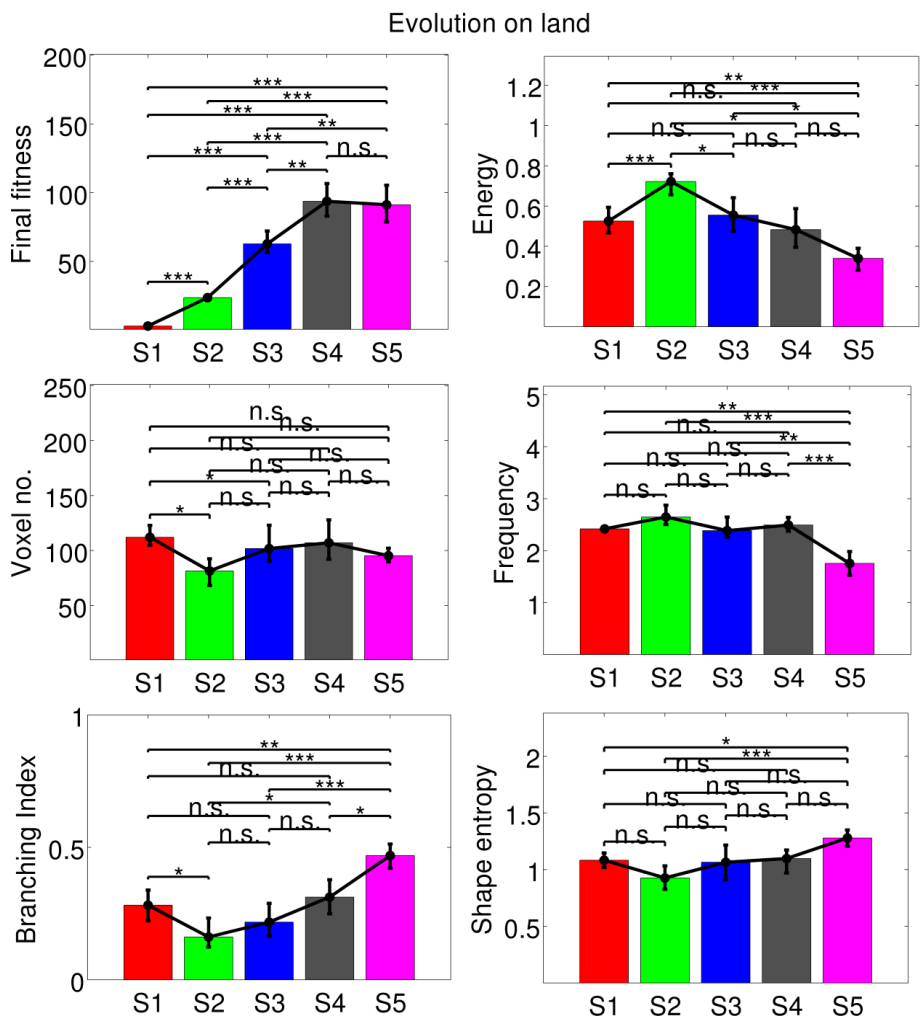

**Figure 7 Evolution on land, descriptors. Final fitness: confirms the trend already noted in Figure 6: stiffer robots perform better. Energy: apart from S1 (non-functional robots), a decreasing trend is observed. Stiffer robots consume less energy. This might be due to the fact that as robots get stiffer, they are able to sustain more effective locomotion strategies (e.g. walking, as opposed to crawling/wiggling around/vibrating). Voxel no.: n.s. Frequency: stiffer robots use lower actuation frequencies: these robots tend to exhibit coordinated walking patterns which can be achieved with slower actuation, while softer ones tend to move by vibrating or bouncing around. Branching index: again, the increasing trend captures the formation of interesting morphological structures, such as appendages to support walking. Shape entropy: a slightly increasing trend confirming what we have noted so far can be noted, although n.s. in most cases.**

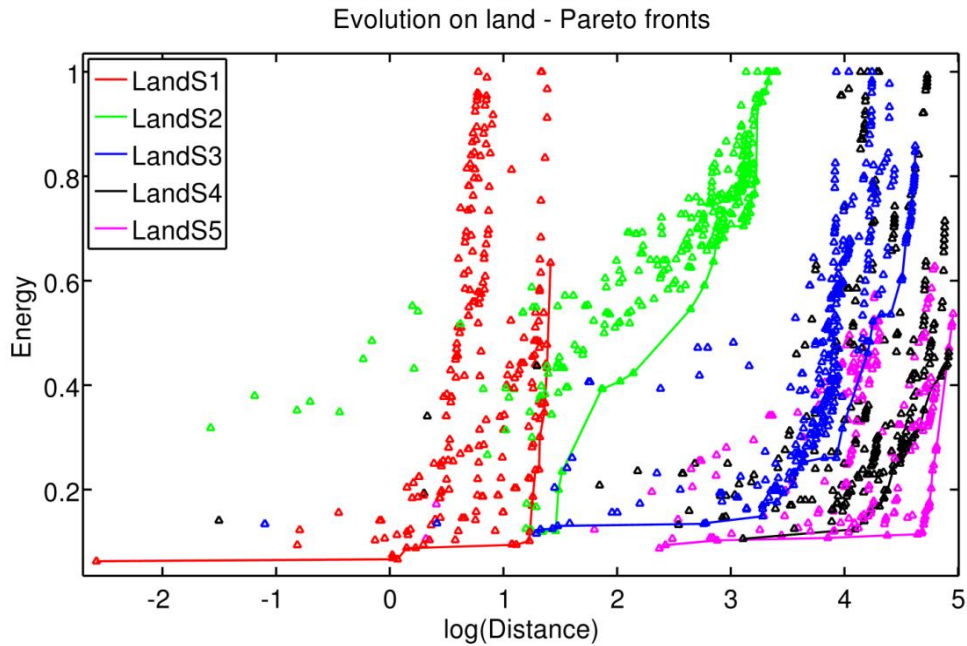

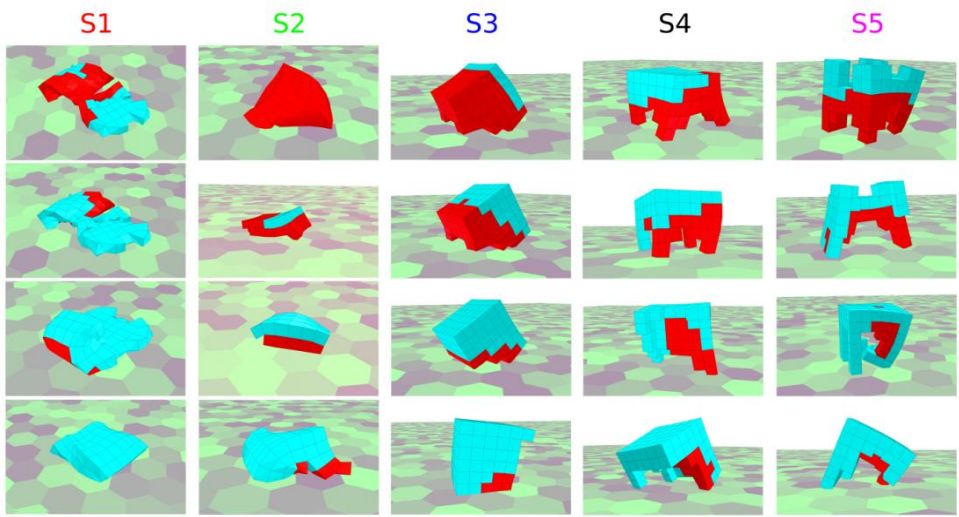

**Figure 8 Evolution on land.** Top: aggregated pareto fronts for runs with different material stiffness. Non-dominated fronts are highlighted with solid lines. The optimum lies in the bottom-right corner of the plot (robots traveling far using very little energy). The plot further highlights that stiffer robots reach better tradeoffs between performances and energy usage. Bottom: some morphologies sampled from the pareto fronts. S1 robots simply collapse under the effect of gravity. S2 robots start to perform some basic form of locomotion, such as crawling and inching. S3 robots start to bounce and run, but best performing robots need a lot of material in order to sustain their weight. With S4 and S5, robots gain an upright posture, and start to develop elaborate walking patterns. Some of the evolved walking robots can be seen in action at: https://goo.gl/vNlzy9

Figure 7 also shows how in many cases evolution places flexible passive tissue on the back of evolved creatures (which serves as support), while limbs are usually composed of active, muscle like material. Figure 10 reports some examples of spontaneous and dynamic transitions from quadrupedal to bipedal locomotion and vice-versa, observed in some of the robots. Finally, it is to be noted that the emergence of highly symmetric and regular patterns is not explicitly rewarded: it is partly enforced by the encoding (CPPNs), partly selected for by the task. During evolution there are many cases in which a final symmetric morphology is reached starting from non-symmetric ancestors.

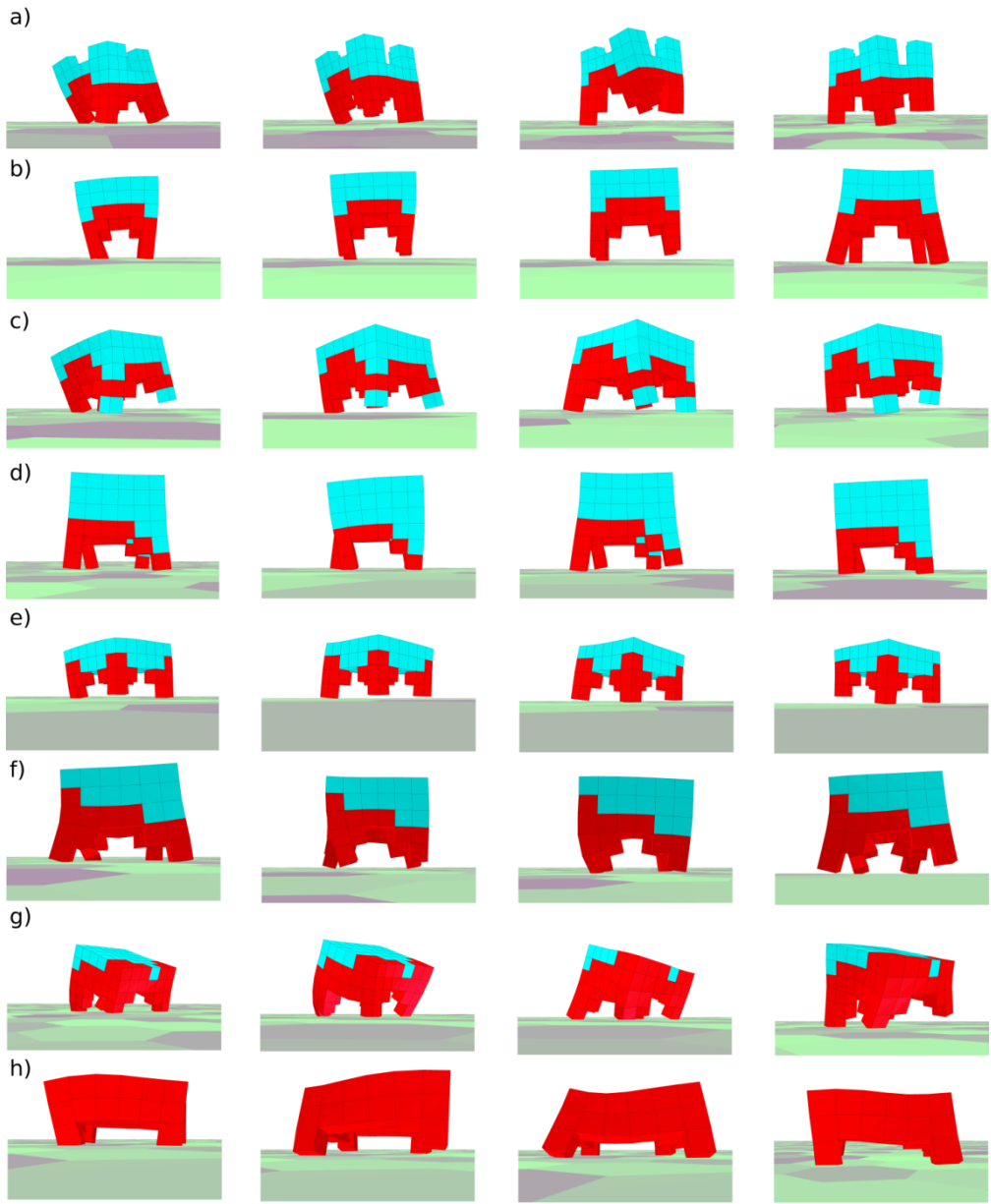

**Figure 9** On land stiffer materials allow the robots to maintain an upright posture which is suitable for walking, as opposed to inching, crawling, bouncing etc. A wide array of upright walking creatures evolve. Here some quadrupeds. Some of these walking robots can be seen in action at: https://goo.gl/vNlzy9

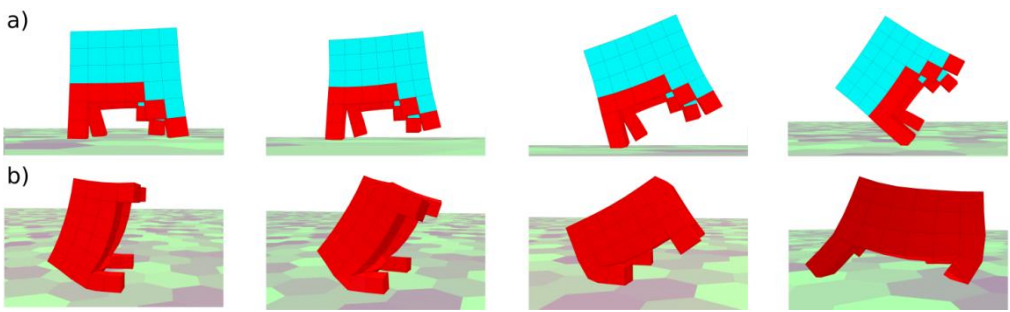

**Figure 10** On land some cases of spontaneous dynamic transition from bipedal to quadrupedal locomotion and viceversa are observed.

## 3.2 Evolution in water

Results of experiments related to evolution in water are reported in this section. Fitness curves are reported in Figure 11, and descriptors in Figure 12. In water an inverse trend is observable with respect to what has been shown for land: the softer the robots, the better the locomotion performances.

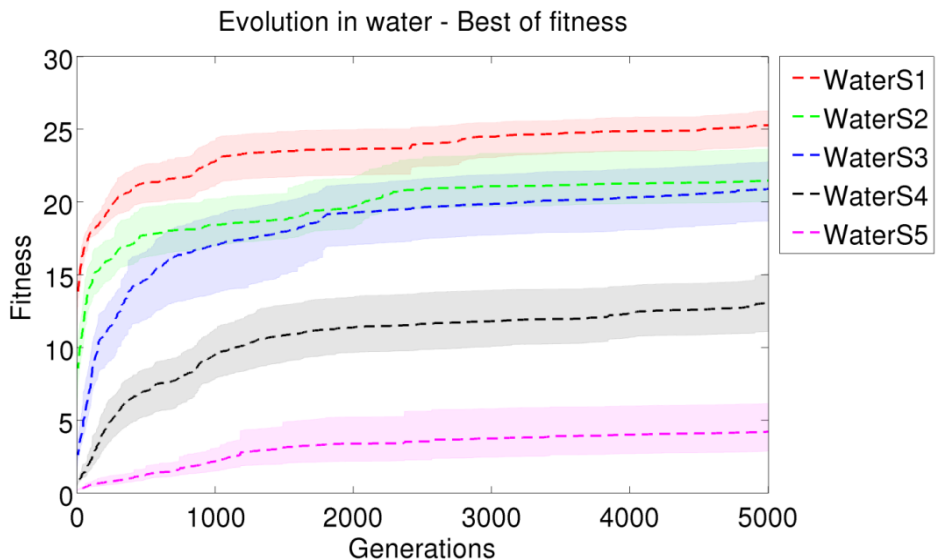

Figure 11 Fitness curves for evolution in water. An inverse trend can be observed with respect to evolution on land. Softer robots (S1) perform better. See Figure 12 for a statistical comparison.

Interestingly, energy efficiency shows a similar trend with respect to land evolution: stiffer robots use less energy (Figure 12). Intuitively, we would have predicted an inverse trend, or a peak of energy efficiency for some intermediate stiffness value, allowing soft robots to move efficiently thanks to the exploitation of passive dynamics. The observed trend does not mean, however, that stiffer robots are more energy efficient. Energy alone does not tell the full story: what is relevant is energy for a given performance level. The aggregate pareto provide some additional insights (Figure 13). Despite softer robots (S1) being on average faster, and stiffer ones (S5) using on average very little energy, there seem to be more complex energy-performances tradeoffs in water. Beyond what can be inferred by the analyses of the descriptors computed on the final populations (Figure 12), the intermediate stiffness S3 actually seems to provide the better energy-performance tradeoffs. This also highlights a difference with respect to evolution on land: while, there, maximum stiffness entailed maximum performances and minimum energy expenditure, in water maximum softness does not represent the optimal choice if we take into account energy expenditure.

Another observable trend in Figure 12 is the one which sees stiffer robots using less material on average. The reason for this is not entirely clear. A possible explanation could be the following one: producing smaller robots in stiffer settings may be a strategy to fight drag resistance in presence of lower swimming forces, generated by robots which cannot exploit passive dynamics to generate high velocities and propulsive forces.

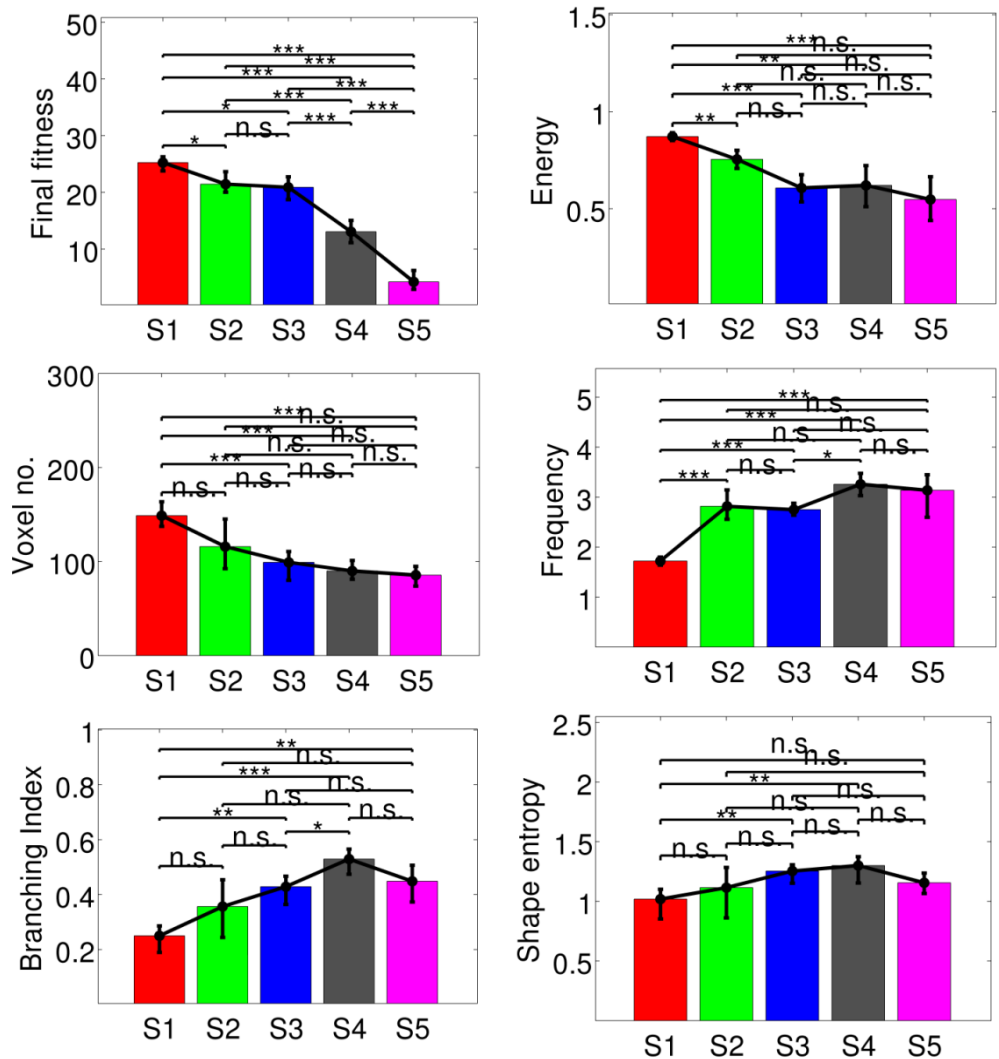

Figure 12 Descriptors for evolution in water. Final fitness: an opposite trend with respect that observed for evolution on land can be noticed. Fitness decreases as material stiffness increases. Energy: as observed on land, stiffer robots tend to use less energy. This does not necessarily correspond, however, to better tradeoffs between performances and energy. Figure 13 is in fact more informative, and tells a different story. Voxel no: stiffer robots tend to use less material. Producing smaller robots may be a strategy to fight drag resistance in presence of lower swimming forces, generated by robots which cannot exploit passive dynamics to generate high velocities. Frequency: differently from evolution on land, stiffer robots seem to use higher actuation frequency here. This may be due to evolution trying to match actuation frequency to the resonant frequencies of freely-moving bodies. Branching index: similar trend w.r.t. land, increases up to S4, then decreases. Need to analyze morphologies. Shape entropy: a slight increase in morphological complexity is noticeable as stiffness increases.

Interestingly, actuation frequency follows an opposite trend with respect to the one observed on land: softer robots use lower actuation frequencies, while stiffer ones use higher ones. This may be due to evolution trying to match actuation frequency to the resonant frequencies of freely-moving bodies [86] [87] [88] [89]. Some of the best performing robots sampled from the aggregate pareto fronts are reported in Figure 13. Most of them developed tentacle-like appendages to propel themselves. Some developed flapping wings. In many cases, evolution discovers that it is possible to have passive appendages while still generating thrust, thanks to the exploitation of passive dynamics.

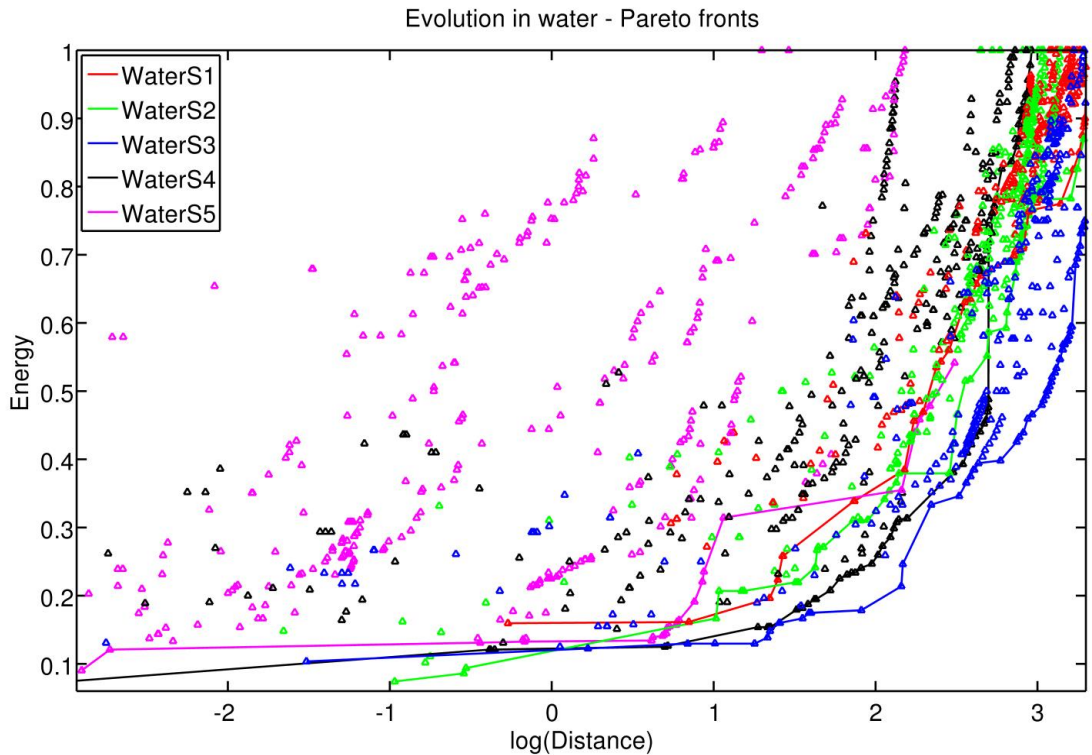

Pareto fronts sampling (rows order: decreasing energy usage):

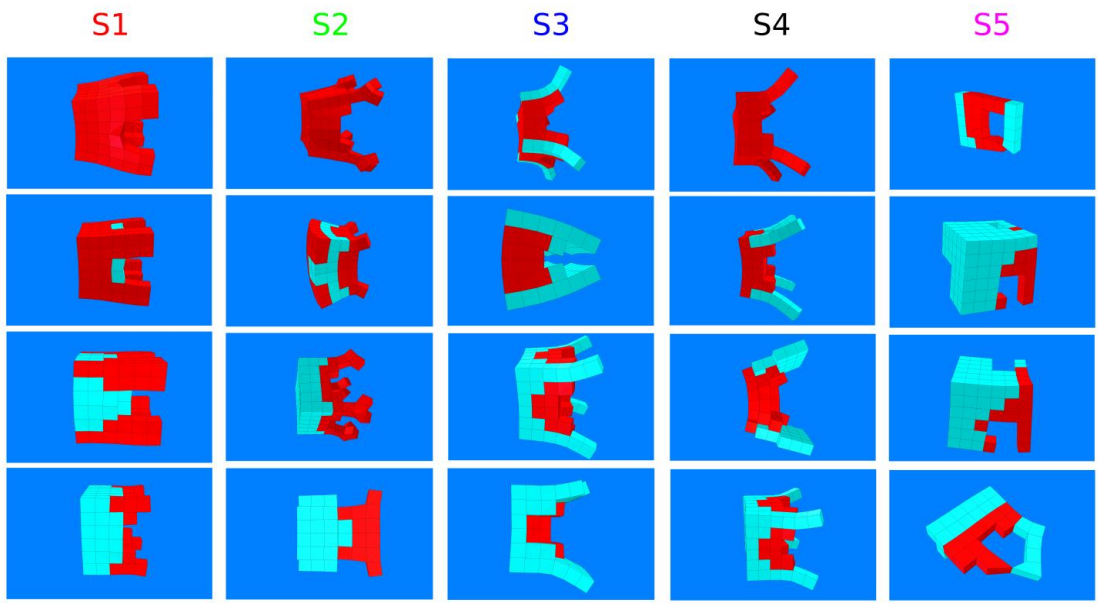

**Figure 13 Evolution in water.** Top: aggregate pareto fronts for runs with different stiffness. Although softer robots perform better than stiffer ones on average (Figure 12), more complex energy-performances tradeoffs are observable here. The best tradeoffs seem to be achieved by the intermediate stiffness S3. Bottom: Some morphologies sampled from the pareto fronts. Most of them developed tentacle-like appendages to propel themselves, which are passive in many cases, exploiting passive dynamics. Some developed flapping wings. In more than one case evolution discovers that it is beneficial to drill a hole in the cross section of the robot in order to minimize drag resistance. Some of the best stiff robots (S5) exploit self-collisions resulting in fast vibrations to produce thrust. Some of these robots can be observed in action at **https://youtu.be/4ZqdvYrZ3ro**

Figure 14 reports instead some of the best performing swimming robots overall. Some of them are hollow inside, featuring a mantle that recalls that of some aquatic creatures (e.g. cephalopods). Differently from these creatures, however, our simplified fluid model does not take into account the thrust generated by the expulsion of ambient fluid due to the compression of this kind of morphological structures. For this reason, some of the fastest creatures (Figure 14a) feature a mantle that is open at both sides, having removed material from the frontal part of the body in order to reduce water resistance on their cross-section. Other robots have flapping wings, others exhibit some kind of undulatory locomotion.

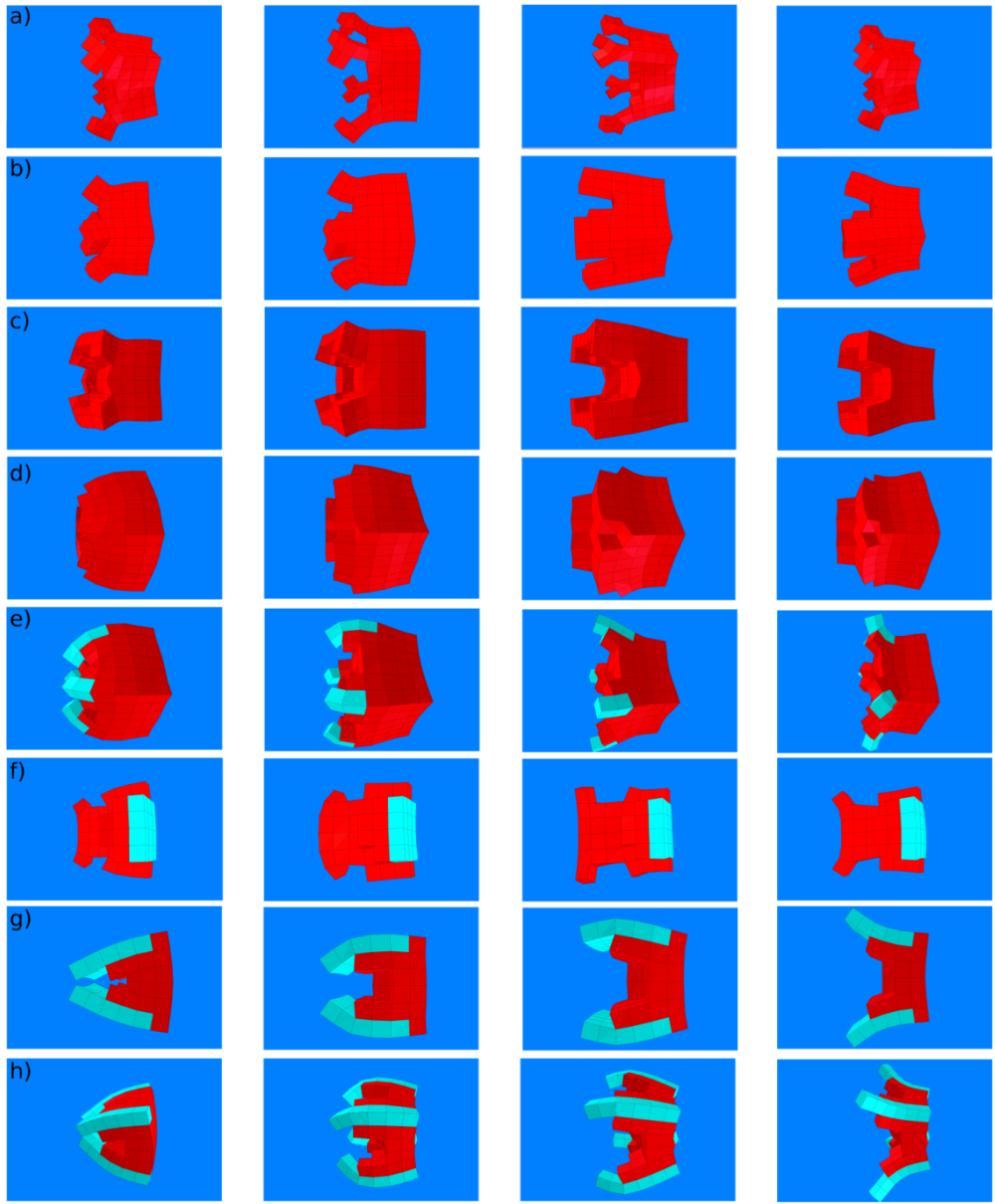

**Figure 14 Some of the best performing swimming robots. Some of them are hollow inside, featuring something like the mantle of some aquatic creatures (e.g. cephalopods). The "mantle" of some of the fastest creatures (a) is instead open at both sides, in order to reduce water resistance on the cross-section (not visible in the picture). Other robots exploit different strategies to move around: (d) performs some kind of undulatory locomotion, (f) pushes water with an oscillating rear piston, (g) bases its locomotion on two large, passive, flapping wings. Some of these evolved creatures can be seen in action at: https://youtu.be/4ZqdvYrZ3ro**

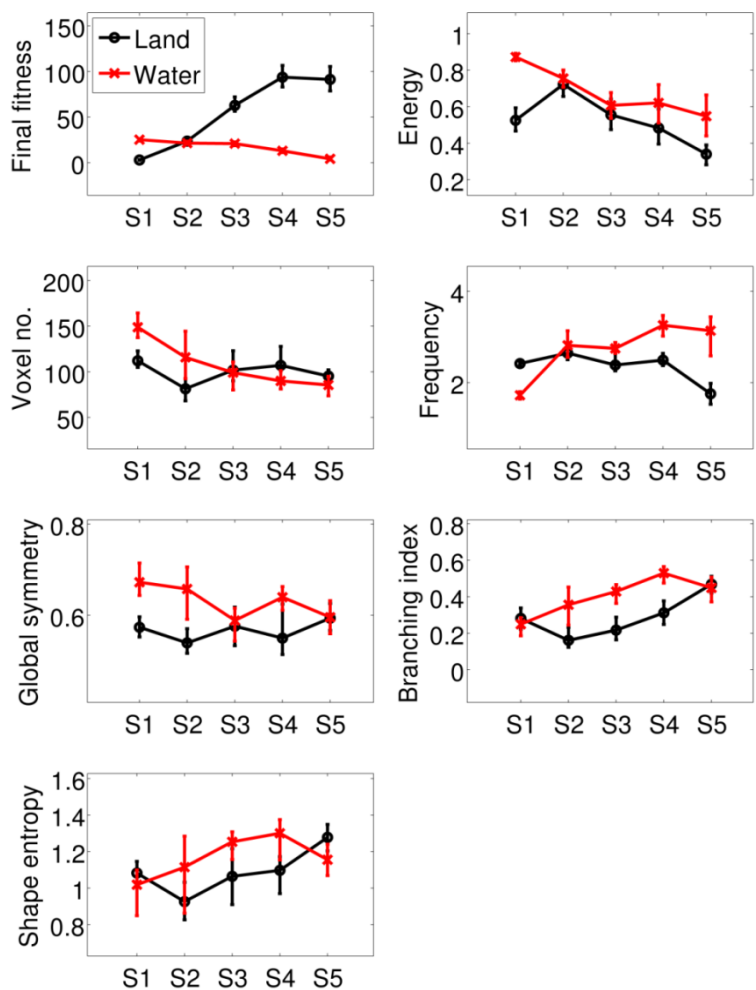

**Figure 15** Comparison among trends in land and water evolution. Some differences and some similarities are observable (discussed in the text).

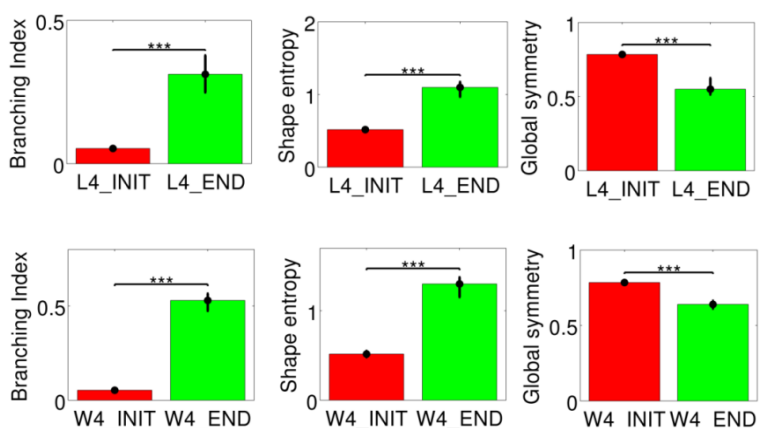

**Figure 16** Comparing morphological descriptors computed at the beginning and at the end of evolution, reported for brevity for the single stiffness S4. Morphological complexity (shape entropy and branching index) spontaneously increases over evolutionary time, without being explicitly rewarded. The marked decrease in global symmetry can be explained by the complexification properties of our generative encoding (a quadruped achieved at the end of the run is less globally symmetric than a full cube, extremely common early on during evolution).

## 3.3 Environmental transitions experiments

### 3.3.1 Water vs Land → Water

The first question to be answered concerns the effect of environmental transitions on evolvability, i.e. whether they yield better or worse results with respect to evolution in a static environment. The answer, for the case of transitions land → water, seems to be clear: transitions land → water seem to be detrimental for swimming evolution (Figure 17).

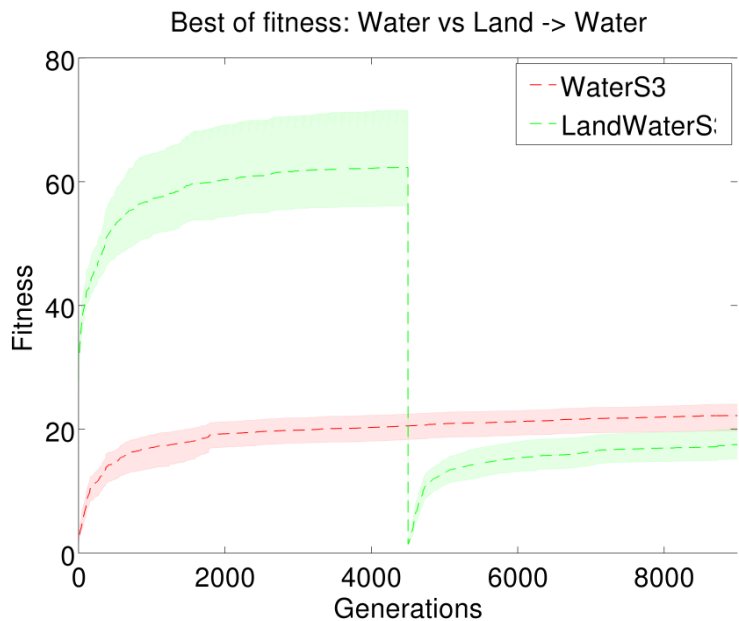

Figure 17 Comparing evolution in water with evolution in presence of the environmental transition land → water. Evolving first on land seems to harm swimming performances ($p < 0.01$). Note how fitness almost drops to zero when the transition is applied, as robots evolved to walk on land start being evaluated for terrestrial walking. The decrease in the absolute value of the fitness in the two environments, instead, purely results from arbitrary choices of parameters associated with friction (land) and fluid drag (water).

The situation appears to be slightly more complex by analyzing the pareto fronts (Figure 18): although evolution on water seems to do better overall, there are some portions of the energy-performance spectrum which are dominated by the experiment where environmental transitions were applied. Interestingly, transition experiments dominate a region of the space which is the closest to the optimal solutions (i.e. maximum locomotion performances with minimum energy expenditure).

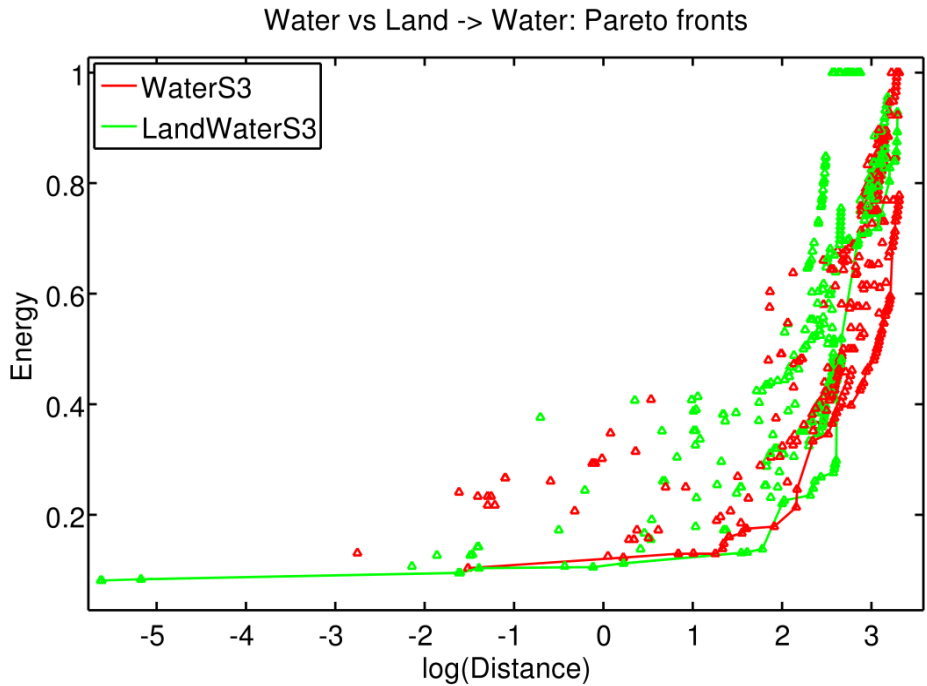

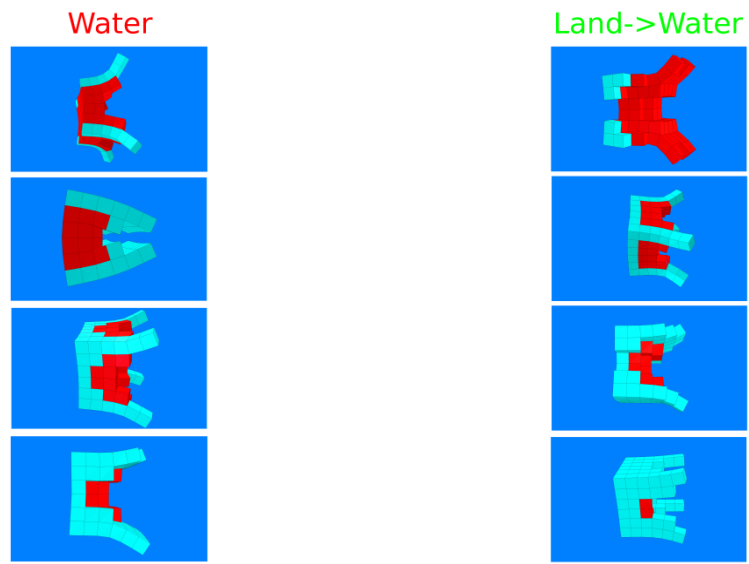

Figure 18 Top: Comparison between aggregate pareto fronts achieved evolving in water and in presence of the environmental transitions land → water. Water experiments dominate most of the energy-performances tradeoffs, although transition experiments do a good job in an interesting region of the search space, close to the optimum (bottom right corner). Bottom: some of the morphologies sampled from the pareto fronts. Note the emergent passive appendages. Some videos of these swimming creatures can be seen in action at https://youtu.be/4ZqdvYrZ3ro

### 3.3.2 Land vs Water → Land

We now analyze the opposite transition, comparing evolution on land with evolution in water followed by a transition to land. For this transition, our hypothesis was that evolving in water first might allow evolution to explore a wider range of morphologies, due to the fact that it may be easier to achieve a smoother fitness gradient in water with respect to land. Especially at the beginning of the run, it may be easier to achieve robots that do not move at all on land, not being able to overcome static friction. In water this does not happen, as robots are free to move and may be able to generate at least some propulsive action. Additionally, we expected some effects of the transition on shape descriptors associated with symmetry and branching, which we expected to be promoted in a water environment. The underlying assumption was that there might be a stronger pressure towards producing appendages in water, which are useful in order to produce relevant speeds and fluid dynamics forces. At the same time, symmetry might be also selected for more easily: with the lack of ground support present on land, it may be easier to move in water for symmetric body plans, it may be easier for regular and symmetric body plans to generate directional forces, instead of just spinning in place. All these aspects might have resulted in more complex and effective morphologies on land.

Figure 19 reports the fitness curves. A promising trend can indeed be observed, with the transition experiment raising above the fitness curve related to evolution on land. The difference, however, is not statistically significant. Technical limitations such as the high variability of results (note the large confidence intervals, denoting a rather weak evolutionary algorithm) and the reduced number of repetitions and generations might be the reason, and we are confident that by overcoming some of these technical limitations we will be able to verify whether there is an underlying regularity we were not able to prove in these experiments.

An interesting result is, however, that there seems to be an asymmetry between land → water and water → land transitions. While land → water was shown to have a negative effect on the average fitness, this is not the case for water → land, which actually suggests an opposite trend.

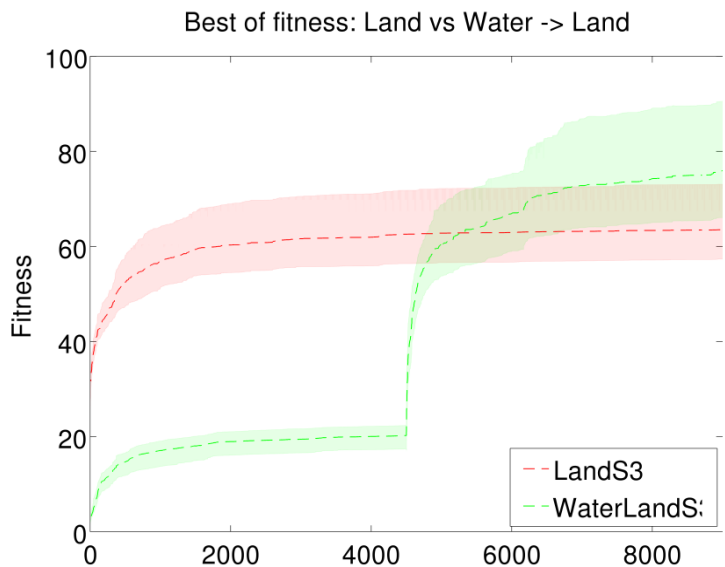

Figure 19 Comparing fitness curves for evolution on land and in presence of the environmental transition water → land. Results indeed show a promising trend which would confirm our hypotheses that see this transition as beneficial, but statistical significance was not achieved (p > 0.05). Technical limitations such as the reduced number of repetitions and generations, as well as a weak evolutionary algorithm (note the large spread of the confidence intervals) might be hindering our hypothesis. An asymmetry is however demonstrated between the two possible types of transitions: while land → water was shown to be detrimental, water → land was not, and actually suggested an opposite trend.

Once again, however, the analysis of the pareto fronts provides additional information. From the analysis of the pareto fronts (Figure 20) it can be noted how the water → land does dominate the whole spectrum of the energy-performance space, which further suggests how this environmental transition may entail some beneficial result. The qualitative morphologies shown in Figure 20 seem to suggest more elaborate solutions being achieved in the water → land experiments, which would confirm some of our hypothesis.

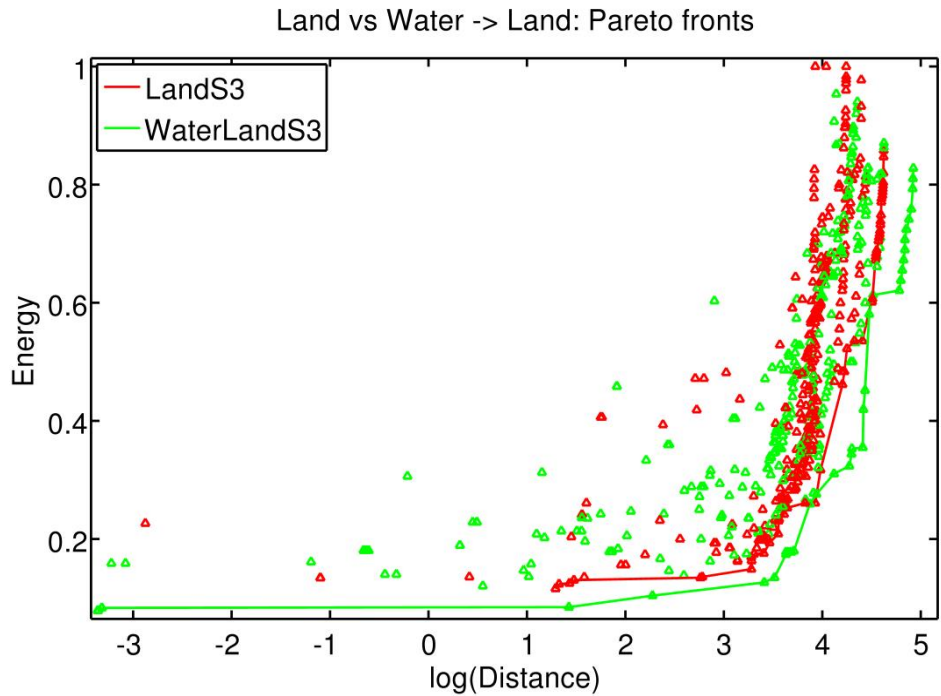

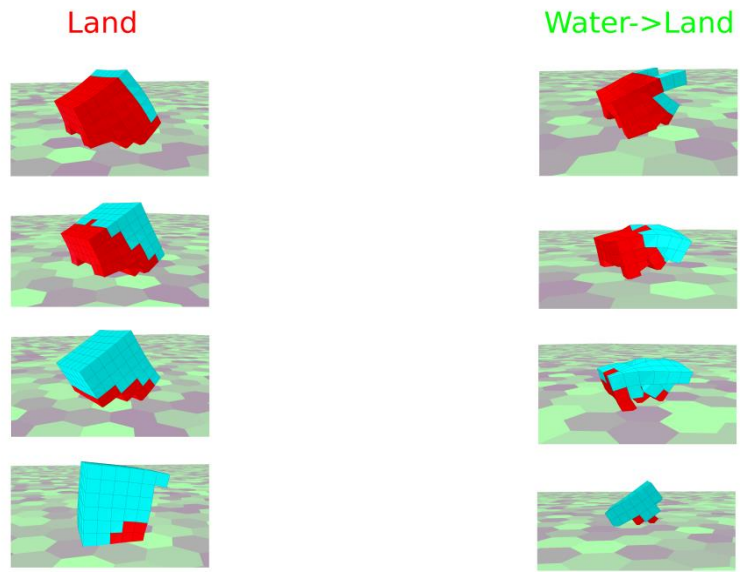

Figure 20 Top: Comparison between pareto fronts achieved in land and in presence of the environmental transition water → land. Although there was no statistical significance in the comparison between the fitness curves, the transition experiment actually produced better results across the whole pareto front, further suggesting the potential beneficial effect of the water → land transition. Bottom: some morphologies sampled from the fronts. Note how the ones achieved in the transition experiment are more elaborate, which is confirmed by trends in Figure 21. The fastest individual can be seen in action here: https://youtu.be/H4D3AXW8yes

From the analysis of descriptors reported in Figure 21 (for both types of transitions), an increase of shape entropy in this kind of transition can indeed be noted, as well as one of the branching index (the latter is not, however, statistically significant). This does not seem to happen for the land → water transition. The only other significant differences entailed by transition experiments are in the evolved actuation frequency, which is higher in both transition experiments, as opposed to those in a static environment. The reason for that is not clear at the moment.

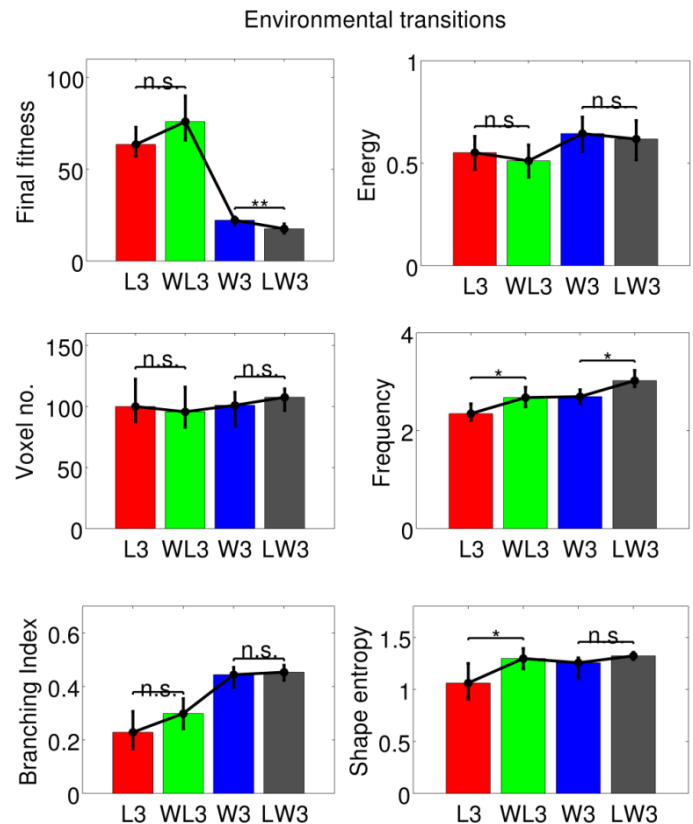

**Figure 21** Comparing descriptors in presence or absence of environmental transitions during evolution (L3: Land S3. WL3: WaterLand S3, etc.). The water → land transition seem to entail an increase in the shape entropy, which agrees with some of our hypotheses. Branching index seems to increase as well, although n.s. A general increase in the actuation frequency is also reported in both transition experiments, its reason not being clear at the moment.

Finally, in Figure 22 some snapshots of different robot phylogenies are reported, which show different types of exaptation phenomena (commented in the figure caption).

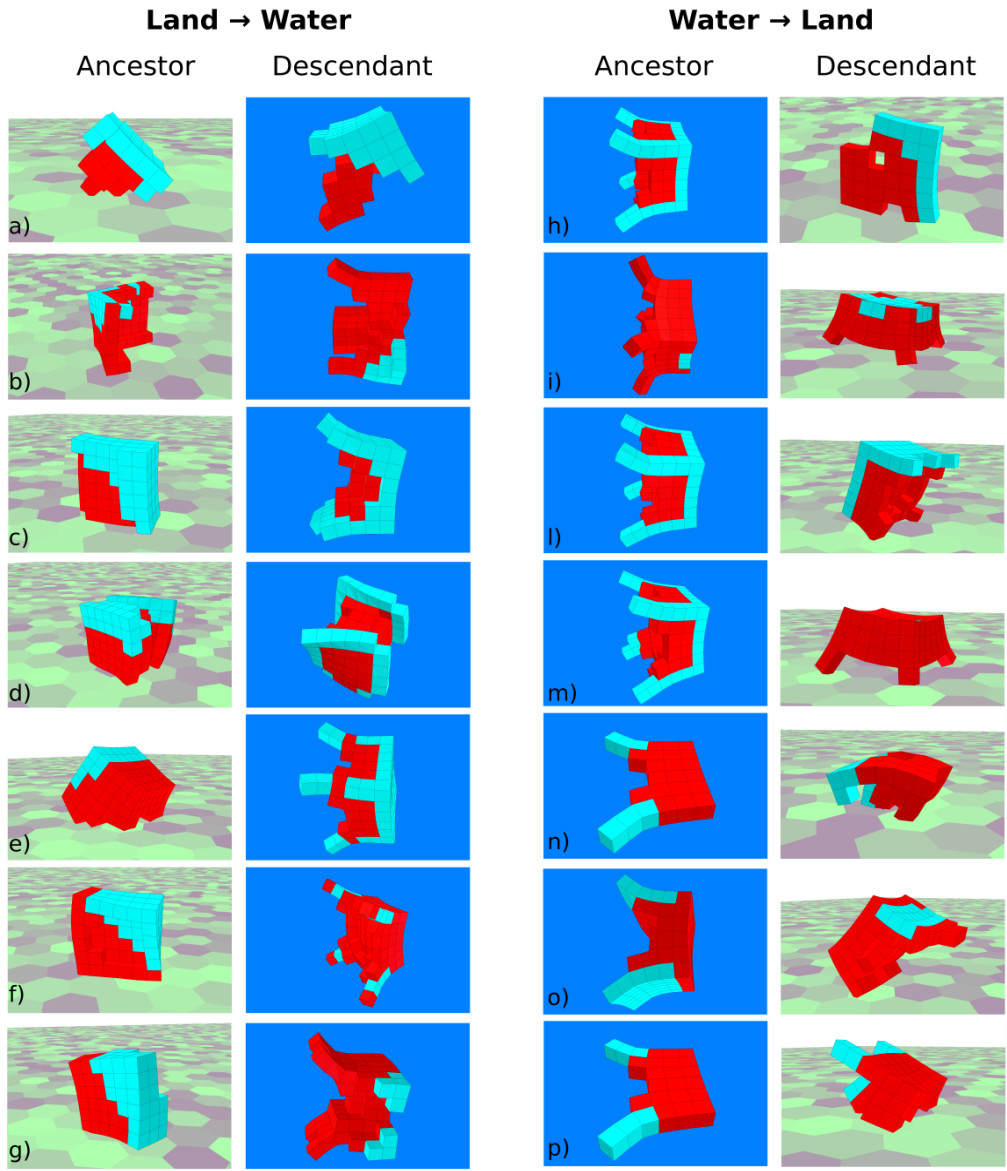

**Figure 22** Some snapshots from robot phylogenies in transition experiments. a) The central part of the body, pushing the robot forward on land, gets longer in water, providing thrust. The uppermost part of the robot gets sharper in water to reduce the resistive drag on the front section. b) A corner-shaped robot evolved to hop on land develops two flapping appendages used to swim in water. c) The basic morphology devised to push on land is adapted and mirrored on both sides in order to achieve a flapping motion in water. d) Here the topology of the robot remains almost identical, with some changes to the actuation patterns and to the distribution of active and passive material, turning a walking robot into a flapping one. e,f,g) When transitioning from land to water, appendages are often developed to facilitate swimming. The opposite phenomenon often happens when moving from water to land (h). More often, appendages that were useful in water are not completely lost when moving onto land, but are instead repurposed. In (i,m) tentacles are shortened, and become legs to support quadrupedal locomotion. In (l,o) slight modifications to the appendages, material distribution and actuation patterns turn flapping robots into hopping ones. In other cases (n,p) robots exploiting flapping tentacles develop additional legs-like appendages on land, as well as a central core used for pushing. The ancestral tentacles visible on the top are not lost, and help maintaining balance on land. Some of these creatures can be seen in action at: https://youtu.be/4ZqdvYrZ3ro

# 4 Conclusion

In this work several regularities have been unveiled regarding how material properties transform the energy-performance tradeoffs found by evolution in different ways in different environments. It has been shown how, on land, locomotion cannot be achieved when evolution is provided with too soft materials. At the same time, increasing material stiffness resulted in more complex morphologies, exhibiting faster, more elaborate, and more energy efficient locomotion. As a result of the wide range of stiffnesses explored in this work, a wealth of quadrupedal morphologies exhibiting clear walking gaits has been evolved in this setup, that were never observed in previous work. An increased level of morphological regularity (clear formation of limbs, symmetry, regular patches of material) is also to be noted with respect to previous works adopting a similar setup (simulator, encoding, optimization algorithm). A possible reason for this could be in the optimization objectives being defined in this work. An opposite relationship between stiffness and locomotion performance is observed in water, with softer robots achieving better locomotion performances than stiffer ones. However, by analyzing the pareto fronts for evolution in water, it is shown how the energy-performance trade-off is more complex in water than it is on land. Particularly, despite softest robots achieving better locomotion performances, it is for an intermediate stiffness level that the best energy-performance tradeoffs are achieved. This tendency to favor structurally-compliant bodies and appendages is consistent, in general, with the observations of aquatic biological systems [90] [87]: flexibility does play a substantial role in enhancing swimming efficiency [89] [91] mainly as a result of fine tuning actuation with the resonant frequencies of the propulsors [92] [88] [86]. However, the marked performance advantage of highly soft organisms predicted by our simulations slightly detaches from actual findings [93], which show how the harmonic oscillations of stiffer appendages in water is often associated with larger thrust [94] [91]. These simulations succeed at tracing some salient features of aquatic locomotion. However, the approximations adopted prevent the model from capturing the dynamics associated with vortex formation, thus precluding the evolution of fish-like creatures. The creatures evolved in this vortex-less and inertialess fluid can only propel themselves by beating their appendages against the surrounding medium as a mean to generate thrust. Having neglected added-mass contributions, pulsed-jetting modes cannot be successfully predicted, thus overlooking squid-like creatures. The outcome consists of organisms vaguely resembling medusoids and morphologically similar among themselves over the stiffness spectrum. The anatomical uniformity of the evolved morphologies is due to the overly-simplistic nature of the fluid-body interaction which constrains the degree of variability of the modes of locomotion and, as a consequence, of the emerging morphologies. Nonetheless, within the evolutionary trend predicted by the model, patterns which are consistent with those of aquatic biological system spontaneously emerge.

Another difference between evolution on land and in water lies in the adopted actuation frequency. On land, actuation frequency decreases as stiffness increases, as we move from inching, crawling ad hopping robot toward walking ones, requiring a slower control signal and carefully tuned phase offsets in order to coordinate their articulated walking motion. In water, instead, actuation frequency increases as stiffness increases, which may be explained by evolution trying to match the resonant frequency of the increasingly stiffer robots freely moving in water. It is shown how by penalizing energy usage, evolution is able to strategically employ passive tissue. This happens both on land and in water, where there are many examples of passive appendages contributing to the generation of fluid dynamics forces. On land, spontaneous dynamic transitions from bipedal to quadrupedal gaits and vice versa are also reported. A general, spontaneous increase in the morphological complexity is noted during evolutionary time, which agrees with previous results [82].

We then started investigating the effect of environmental transitions on the evolution of morphologies and behaviors. An asymmetry is reported concerning the effect of transitions water → land: while moving from land to water resulted to be detrimental for the evolution of swimming, the opposite transition (water → land) pointed out some benefits for the evolution of walking, yielding more optimal solutions in the energy-performance space and more elaborate morphologies. Also, water → land environmental transitions seem to produce more elaborate morphologies. Finally, by analyzing some of the phylogenies in the transition experiments, several examples of exaptation phenomena are commented, which are shown to happen in an artificial scenario in analogy to what is observed in biology. When moving from land to water appendages are produced, which are useful in order to generate swimming forces. Conversely, appendages are often lost, shortened, or re-purposed as robots are moved from water to land. Although pointing out interesting phenomena and regularities, the transition experiment do not bring in some cases definitive evidences, which may be related to some technical limitations. Particularly, they were performed for a single stiffness value (and we have showed how this parameter dramatically affects results). Experiments would have to be repeated for different stiffness values, and observations compared. A specific type of transition may be more or less beneficial depending on the particular material stiffness value. In general, in transitions experiments as well as in those performed in static environment, it would be interesting to evolve the stiffness distribution. This feature has been implemented already in our setting, and preliminary experiments performed, which are not reported here for brevity. Other observed limitations are related to the evolutionary algorithm, that appears to be rather weak, as can be noted by the considerable spread of the fitness curves. This makes very difficult to achieve statistical significance when comparing different treatments. We are currently investigating the use of more robust evolutionary algorithms (such as the Age-Fitness Pareto Optimization [95]) in order to overcome some of these limitations. Other limitations are mainly due to time constraints, such as the reduced number of repetitions for each treatment, as well as the number of generations, which could be higher for transition experiments. Finally, an obvious limitation lies in the adopted fluid model, which is overly simple for computational reasons. All of our observation need to be carefully analyzed and compared with evidences from robotic and fluid dynamics studies, and are therefore limited, for the time being, to our simulated setup, offering a world dominated by resistive fluid drag only without turbulent phenomena whatsoever. More realistic fluid models will also require the possibility to evolve more complex actuation signals, beyond the simple harmonic oscillators currently implemented. A number of aquatic biological systems are, in fact, found to rely on swimming cycles where impulsive thrusting phases are associated with ramp down, recovering ones, which helps inducing non-symmetric inertial effects which result in a positive net thrust. This is the case of certain cephalopods or medusoids, but examples are widespread across different species. This suggests that non-harmonic actuation routines are of importance in unsteady aquatic locomotion, and should be taken into account.

# 5 Acknowledgments

This work was supported by National Science Foundation awards INSPIRE-1344227 and PECASE-0953837, and the United States Army Research Office award W911NF-16-1-0304. F. Corucci is supported by grant agreement no. 604102 (Human Brain Project) funded by the European Union Seventh Framework Programme (FP7/2007-2013). N. Cheney is supported by NASA Space Technology Research Fellowship #NNX13AL37H. F. Giorgio-Serchi is supported by the Natural Environment Research Council (grant number NE/P003966/1). Computational resources were provided by the Vermont Advanced Computing Core (VACC).

# 6 Author Disclosure Statement

No competing financial interests exist.